\documentclass[final]{cvpr}

\usepackage{times}
\usepackage{epsfig}
\usepackage{graphicx}
\usepackage{amsmath}
\usepackage{amssymb}
\usepackage{xcolor}
\usepackage{multirow}
\usepackage{multicol}
\usepackage{booktabs}
\usepackage{subcaption}

\newcommand{\iie}{\emph{i.e.}}

\usepackage[pagebackref=true,breaklinks=true,colorlinks,bookmarks=false]{hyperref}

\def\confYear{CVPR 2023}

\begin{document}

\title{DOAD: Decoupled One Stage Action Detection Network}

\author{Shuning Chang\textsuperscript{\rm 1}\thanks{Work done during an internship at Alibaba Group.}\quad Pichao Wang\textsuperscript{\rm 3}\thanks{Equal corresponding authors.}\quad Fan Wang\textsuperscript{\rm 3}\quad Jiashi Feng\textsuperscript{\rm 2} \quad Mike Zheng Shou\textsuperscript{\rm 1}\footnotemark[2] \\ 
\textsuperscript{\rm 1}Showlab, National University of Singapore \quad \textsuperscript{\rm 2}National University of Singapore \quad \textsuperscript{\rm 3}Alibaba Group \\
}

\maketitle

\begin{abstract}
Localizing people and recognizing their actions from videos is a challenging task towards high-level video understanding. Existing methods are mostly two-stage based, with one stage for person bounding box generation and the other stage for action recognition. However, such two-stage methods are generally with low efficiency. We observe that directly unifying detection and action recognition normally suffers from (i) inferior learning  due to different desired properties of context representation for detection and action recognition; (ii) optimization difficulty with insufficient training data. In this work, we present a decoupled one-stage network dubbed DOAD, to mitigate above issues and improve the efficiency for spatio-temporal action detection. To achieve it, we decouple detection and action recognition into two branches. Specifically, one branch focuses on detection representation for actor detection, and the other one for action recognition. For the action branch, we design a transformer-based module (TransPC) to model pairwise relationships between people and context. Different from commonly used vector-based dot product in self-attention, it is built upon a novel matrix-based key and value for Hadamard attention to model person-context information. It not only exploits relationships between person pairs but also takes into account context and relative position information. The results on AVA and UCF101-24 datasets show that our method is competitive with two-stage state-of-the-art methods with significant efficiency improvement.
\end{abstract}
\vspace{-4mm}

\section{Introduction}
The objective of action detection is to localize and recognize human actions in video clips along space and time. Unlike general action recognition, the actions in this task emphasize on actors' interactions with the context. As a fundamental and challenging task in video understanding, it has been widely applied to various tasks, such as abnormal behavior detection~\cite{xu2015learning,lee2019bman} and autonomous driving~\cite{DBLP:journals/corr/abs-1912-00438}.

\begin{figure}[t]
\includegraphics[width=1.0\linewidth]{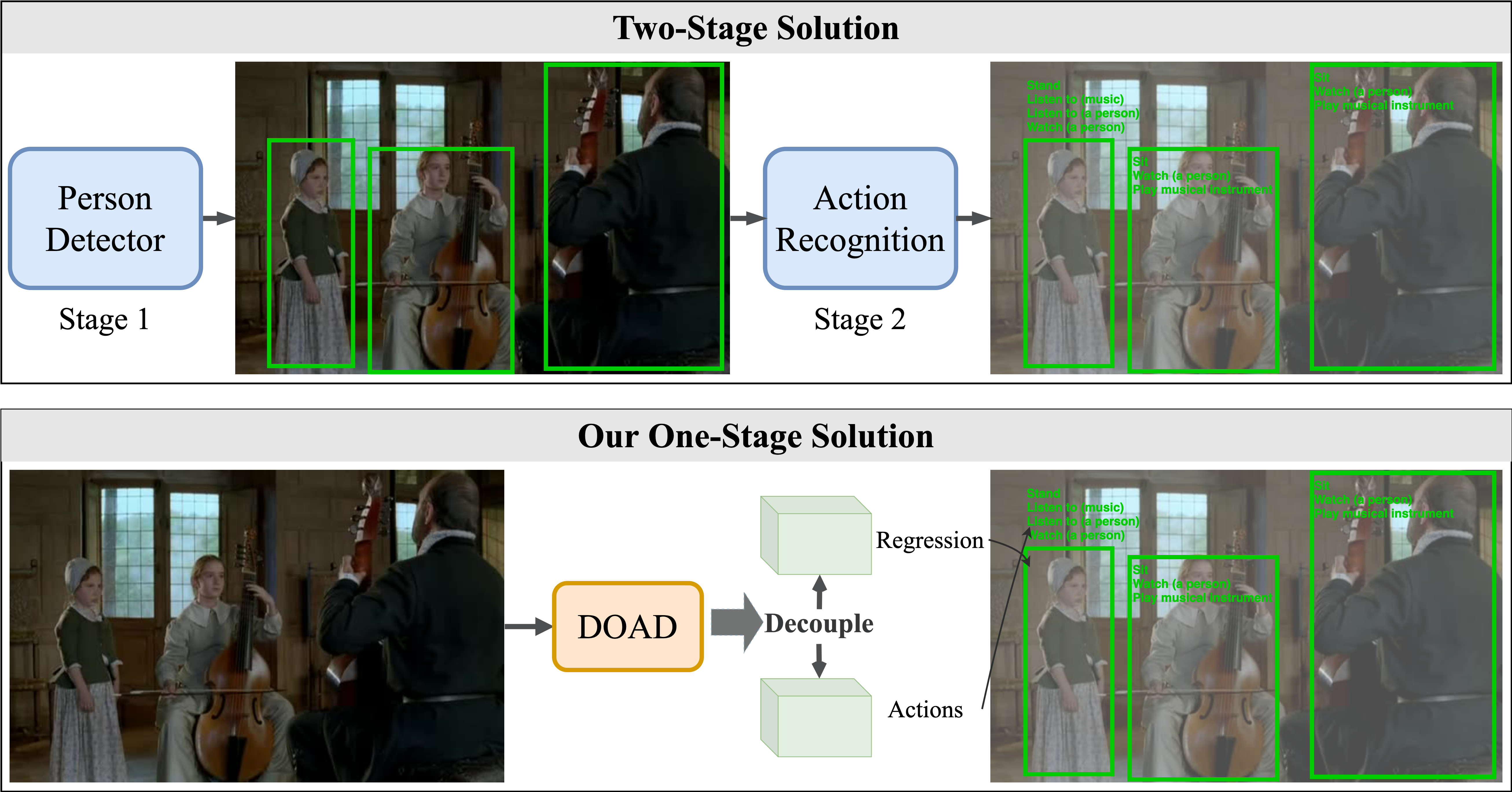}
\centering
\caption{Comparison between two-stage solution and our one-stage solution to spatio-temporal action detection. Traditional two-stage methods use an off-the-shelf detector to generate person bounding boxes suffering low efficiency. The proposed DOAD model decouples detection representation and action recognition representation to accomplish different sub-tasks in a single stage.}
\label{first image}
\vspace{-5mm}
\end{figure}

\begin{figure}[t]
\begin{subfigure}{.5\textwidth}
\includegraphics[width=0.9\linewidth]{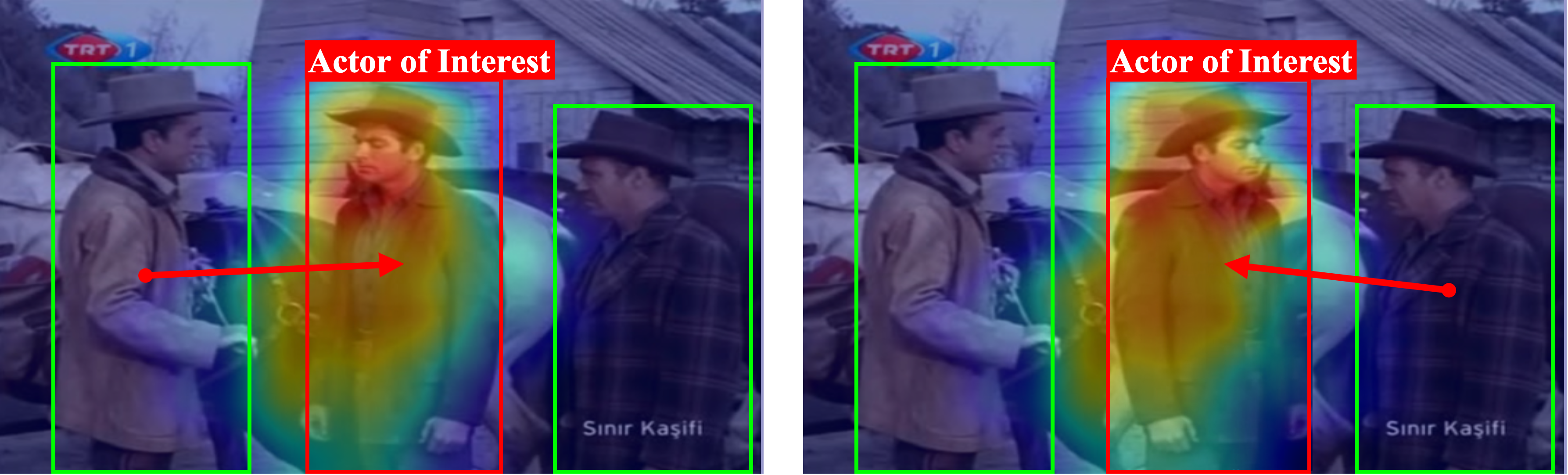}
\centering
\caption{Different context support for detection and action recognition. }
\label{figure2:a}
\end{subfigure}
\begin{subfigure}{.5\textwidth}
\includegraphics[width=0.9\linewidth]{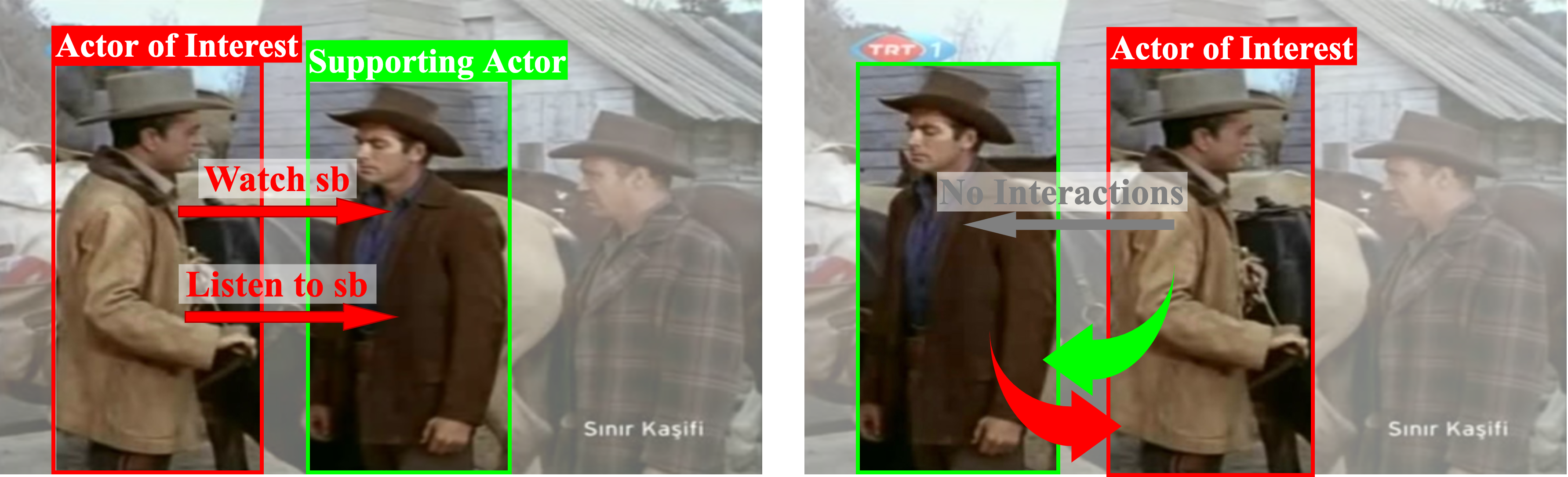}
\centering
\caption{The importance of the context and position information for action recognition. }
\vspace{-2mm}
\label{figure2:b}
\end{subfigure}
\centering
\caption{
(a) shows different context support for person detection and action recognition. The heatmap represents the informative context for detecting actor-of-interest. Changing his pose from the left image to the right image does not change the context for detection, but does change the context (supporting person) for action recognition to another person in the image.
(b) illustrates that the context and position information are crucial clues for action recognition. In the right image, we exchange the positions of two people, then they have no interaction.
}
\vspace{-5mm}
\label{figure2}
\end{figure}

Spatio-temporal action detection usually consists of two sub-tasks, person detection and action recognition.
Most existing methods typically adopt two-stage solutions. As shown in Figure \ref{first image}, they typically follow the top-down strategy~\cite{DBLP:conf/cvpr/GuSRVPLVTRSSM18,DBLP:conf/eccv/SunSVMSS18,DBLP:conf/cvpr/YangY0XDK19,Wu_2019_CVPR,DBLP:conf/eccv/WuKWZW20,DBLP:conf/eccv/TangXMPL20,DBLP:conf/cvpr/PanCSLS021} that employs off-the-shelf detectors to localize person instances at first and then recognize their action categories with various backbones. Though with high performance, these methods are not efficient as they require two-stage processing for detection and action recognition separately.
We observe that detection and action recognition have different desired properties for context representation in this task. Due to the existence of interaction categories, action recognition requires to fuse corresponding entity (\eg, other people or objects) features from context to construct various interaction relationships. In contrast, detection also benefits from context, but it tends to incorporate the context features that support the bounding box regression and is not sensitive to interaction. Take Figure \ref{figure2:a} as a toy example. We slightly change the pose of the interested person from the left image to the right image. The aggregated context features for detection are nearly unchanged, but for action recognition they shall be learned from scratch again due to the change of interaction objects.
The different objectives of the two sub-tasks need different context supports for optimal learning.
Moreover, entangled modeling of the two sub-tasks leads to difficulty in optimization, especially in the case of limited video data with annotations.
Due to these reasons, it is difficult to integrate them into entangled one-stage framework to achieve strong performance.

To alleviate these issues, different from previous methods, we propose a novel decoupled one-stage architecture to unify detection and action recognition into one backbone. Our method decouples detection and action recognition into two separate branches to make the two sub-tasks learn their own optimal context support. 
In our architecture, person detection is performed in the detection branch, which aggregates context information from temporal supporting proposals, and action recognition is performed in the action branch for person-person or person-context relationship mining.

Specifically, in the detection branch, a region proposal network (RPN) is adopted to generate person bounding boxes, and an ROI pooled person feature is served for bounding box regression. Considering the detection is performed on video frames, we employ a temporal aggregation module to enhance the keyframe features by aggregating its adjacent frame features. In the action branch, we design a transformer-based structure which unifies person and context features to capture interaction representation. In this task, action recognition shall understand actors' interactions with surrounding context, including environments, other actors, and objects. Prior works~\cite{DBLP:conf/eccv/WuKWZW20,DBLP:conf/eccv/TangXMPL20} focus on building various interaction relationships, such as person-person, person-object, and long-term temporal interactions. However, they just model corresponding entity features in each relationship and stack them independently, which neglects the correlation among person features,  context and relative position information. Take Figure \ref{figure2:b} as an example. There are two person-person interaction categories, ``watch sb" and ``listen to sb", between the actor-of-interest in the red bounding box and the supporting actor in the green bounding box. 
Assuming the two actors exchange their positions, their appearance features are nearly unchanged, but there are no interactions between them any more. If we do not consider the context and position information, wrong results might be obtained. Therefore, we argue that the relationships among people, context, and position shall be considered simultaneously. Inspired by the vanilla transformer~\cite{vaswani2017attention}, we design a TransPC (Transformer for Person-Context) layer  which models pairwise person relationships upon their holistic spatial structure to retain the context and relative position information. 
We construct matrices of pairwise person-context features as key and value. Different from  vanilla self-attention that deals with sequence input, we propose the Hadamard product to compute attention map between the sequence query and the matrix key. Our TransPC is able to incorporate features from more informative entities and produce more accurate action prediction.

Our contributions are summarized as three-fold:
\begin{itemize}
\item We propose a one-stage spatio-temporal action detection model, which decouples detection representation and action representation for person detection and action recognition, respectively, ensuring that they have optimal context feature aggregation.
\vspace{-1mm}
\item We propose a novel transformer-based method, TransPC, to explicitly integrate person and context features with relative position information for action recognition.
\vspace{-1mm}
\item We demonstrate the effectiveness of our method on the mainstream datasets, AVA and UCF101-24. Our method outperforms well established state-of-the-art one-stage methods significantly and is comparable to the two-stage methods with significant improvement of efficiency.
\vspace{-3mm}
\end{itemize}


\section{Related work}
\paragraph{Action recognition.}
Action recognition is a fundamental task of video understanding.
Convolutional networks have long been the standard for this task.
They can be roughly separated into two groups, \iie, two-stream networks and 3D CNNs. 
Two stream networks~\cite{simonyan2014two,feichtenhofer2016convolutional,wang2016temporal,wu2019convolutional} use 2D CNNs to extract frame features from RGB and optical flow sequences, while 
3D CNNs~\cite{tran2015learning,carreira2017quo,qiu2017learning,feichtenhofer2019slowfast,li2020spatio} adopt 3D convolutional layers to model the temporal information from the original videos. 
Since 3D convolutional networks consume more computation, many methods explore to decouple spatial and temporal dimensions or use grouped convolutions~\cite{sun2015human,tran2019video,tran2018closer,xie2018rethinking,feichtenhofer2020x3d}. 
With the significant success of Vision Transformer (ViT)~\cite{dosovitskiy2020image}, a shift in action recognition from CNNs to transformers emerges recently. Benefit from the self-attention mechanism which broadens wider receptive field with fewer parameters and lower computation costs, those methods~\cite{DBLP:conf/icml/BertasiusWT21,DBLP:journals/corr/abs-2103-15691,DBLP:journals/corr/abs-2106-13230,DBLP:journals/corr/abs-2104-11227,DBLP:journals/corr/abs-2108-11575,DBLP:journals/corr/abs-2106-11297} present state-of-the-art performance and impressive potential.
In this work, besides traditional CNN-based networks, we also try to adopt transformer-based networks as our backbone to extract video features.
\vspace{-5mm}

\paragraph{Spatio-temporal action detection.}
Action recognition processes well-trimmed videos, where the models only need to classify short video clips to action labels. However, most videos are untrimmed and long in practical applications. Recent works explored temporal action localization~\cite{shou2016temporal,zhao2017temporal,chao2018rethinking,lin2019bmn,bai2020boundary,zhao2020bottom,DBLP:journals/corr/abs-2103-16024} and spatio-temporal action detection~\cite{Girdhar_2019_CVPR,DBLP:conf/cvpr/GuSRVPLVTRSSM18,DBLP:conf/eccv/SunSVMSS18,DBLP:conf/cvpr/YangY0XDK19,kopuklu2019yowo,Wu_2019_CVPR,DBLP:conf/eccv/WuKWZW20,DBLP:conf/eccv/TangXMPL20,DBLP:conf/cvpr/PanCSLS021} on untrimmed videos. Spatio-temporal action detection is more difficult than action classification and temporal action detection because these models need to not only predict the action categories but also localize the action in time and space. Most recent works focus on capturing various interaction relationships between actors and context. They normally adopt a two-stage framework where actor boxes are first generated by an off-the-shelf detector and then classified. Wu et al.~\cite{DBLP:conf/eccv/WuKWZW20} present a strong baseline network by simply expanding actor bounding boxes and incorporating global feature, which demonstrates the importance of the context information. Tang et al.~\cite{DBLP:conf/eccv/TangXMPL20} explore nearly all the main interactions including person-person, person-object, and long-term temporal interaction. They model each interaction by the self-attention mechanism and then stack them to improve the performance. Moreover, an Asynchronous Memory Update (AMU) algorithm is proposed to estimate the memory features dynamically for long-term temporal interaction capture. Pan et al.~\cite{DBLP:conf/cvpr/PanCSLS021} show an Actor-Context-Actor Relation model to uncover the deeper level relationship between actors and context by applying a high-order relation reasoning to build the actor-context-actor relations. Those methods achieve significant performance. However, two-stage approaches are not efficient, which limits their application in the real world. Girdhar et al.~\cite{Girdhar_2019_CVPR} adopt an RPN network to generate bounding box proposals and use a transformer head to generate classification and bounding box regression results. Their method is a one-stage method but lacks finer interaction relationship construction and does not consider the difference of optimal context support for two sub-tasks. In this work, we propose a one-stage method that contains our novel person-context interaction relationship mining module and addresses these issues. Experiments demonstrate that our method outperforms competitive baselines.

\paragraph{Attention mechanism for video context capture.}
Context information capture plays a pivotal role in video understanding. Attention mechanism is one of the most effective and common technique to solve it. Attention mechanism is to compute the response at a position in a sequence by accessing all positions and taking their weighted average in the embedding space. Vaswani et al.~\cite{vaswani2017attention}
first introduce a self-attention mechanism, called transformer, capturing long-range context among words in one sentence to address the machine translation task. Girdhar et al.~\cite{Girdhar_2019_CVPR} first utilize transformer in video tasks to aggregate features from the spatio-temporal context for recognizing and localizing human actions; after that, many related works~\cite{zhou2018end,seong2019video,wang2021transformer} apply transformer in various video tasks.
There are some other ways besides transformers to utilize attention mechanism to capture long-range dependencies.
Wang et al.~\cite{Wang_2018_CVPR} embed non-local structure into the action recognition network to capture spatio-temporal context dependencies.
Wu et al.~\cite{Wu_2019_CVPR} introduce long-term temporal context feature banks to compute interactions between the current short-term features and the long-term features for video analysis. Recent works~\cite{DBLP:conf/eccv/WuKWZW20,DBLP:conf/eccv/TangXMPL20,DBLP:conf/cvpr/PanCSLS021} explore the interactions between people and all kinds of context in this task, \emph{e.g.}, person-person and person-scene. However, most works model each context independently and are short of exploring the associations between different contexts. Although ~\cite{DBLP:conf/cvpr/PanCSLS021} proposes the concept of actor-context-actor relation, we still think it does not pay attention to the relative position relationship of different entities.

\begin{figure*}[t]
\includegraphics[width=1.0\linewidth]{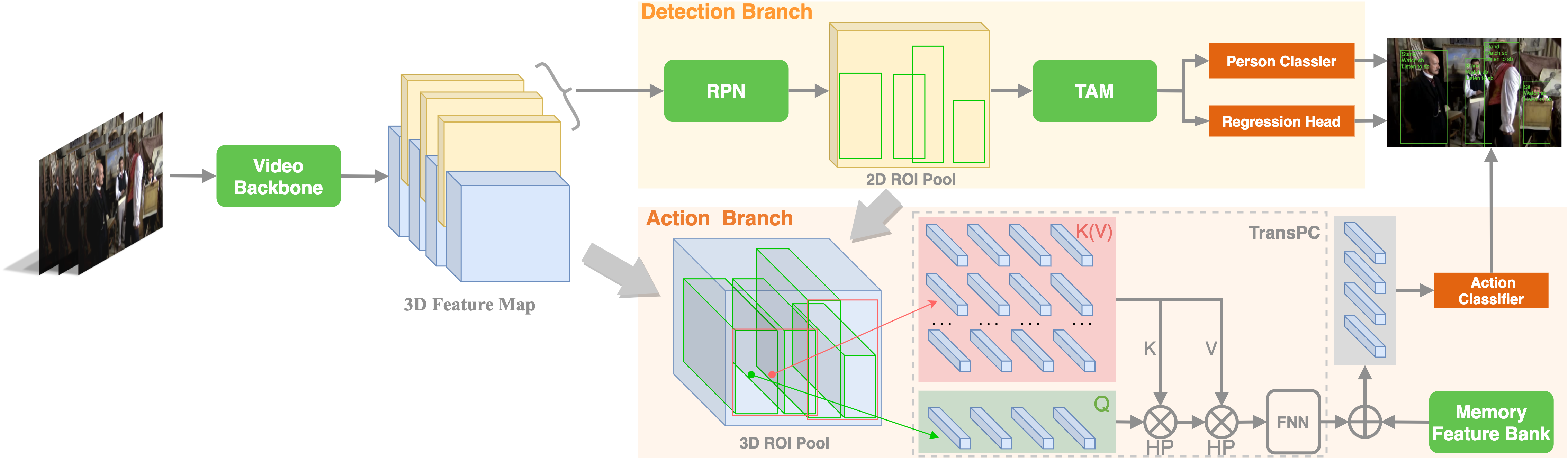}
\centering
\caption{Illustration on the architecture of our method. It first employs a video backbone network to extract a 3D video feature. Then the detection and action recognition is decoupled into two branches. In the detection branch, we apply a similar Faster-RCNN framework incorporating video temporal context by temporal aggregation module (TAM) to generate bounding boxes and conduct pose action estimation. In the action branch, we adopt a TransPC to integrate person and context features to capture interaction relationships. HP is short for Hadamard product.}
\vspace{-3mm}
\label{figure3}
\end{figure*}

\section{Method}
In this section, we present the proposed method which targets to construct an effective and efficient one-stage model. 
The overall architecture of our method is shown in Figure \ref{figure3}. We first employ a video backbone to extract features of the input video. Then, the video representation is decoupled into two branches, \iie, detection branch and action branch. The detection branch (Section \ref{3.2}) is based on Faster-RCNN structure and deploys a temporal aggregation module to generate person bounding boxes. The action branch (Section \ref{3.3}) mines person-context interaction by TransPC module and long-term temporal interaction by a memory feature bank to achieve action recognition. 

\subsection{Overall framework}
\label{3.1}
Our method deals with a short video clip centered on the center frame $\mathcal{F}_k$ (``keyframe"). Following the pipeline of previous spatio-temporal action detection methods~\cite{Girdhar_2019_CVPR,DBLP:conf/cvpr/GuSRVPLVTRSSM18,DBLP:conf/eccv/SunSVMSS18,DBLP:conf/cvpr/YangY0XDK19,Wu_2019_CVPR,DBLP:conf/eccv/WuKWZW20,DBLP:conf/eccv/TangXMPL20,DBLP:conf/cvpr/PanCSLS021}, it generates a set of person bounding boxes for all the people in the keyframe, and recognizes all the actions for each person in this short time. 

We begin by extracting a $T$-frame short clip centered on the given keyframe $\mathcal{F}_k$ from a video. We encode this input using a backbone network (\emph{e.g.} Video Swin Transformer~\cite{DBLP:journals/corr/abs-2106-13230} or SlowFast~\cite{feichtenhofer2019slowfast}) to obtain a $T\times H\times W$ feature map $X$.  
We feed the feature map into the detection and action branches. The detection branch aggregates adjacent frame features to enhance the keyframe person features to generate person bounding boxes. For the action branch, we propose a TransPC module combining person and context features to capture interaction relationships. Finally, we obtain the generated bounding box coordinates and action categories as our final action detection results. We will illustrate our detection branch and 
action branch in Section \ref{3.2} and Section \ref{3.3}, respectively.

\subsection{Detection branch}
\label{3.2}
Our detection branch is similar to the Faster R-CNN object detection framework~\cite{DBLP:journals/pami/RenHG017}. We slice out the keyframe feature $X_k\in \mathbb{R}^{H\times W}$ from feature map $X$ and feed it into a region proposal network (RPN). The RPN generates person bounding box proposals with scores. We then select $N$ proposals $(N=300)$ according to the interaction of overlap (IoU) with ground truths. After that, person features are extracted by align ROI pooling operation from the selected proposals. These features are applied to classify proposals into \textit{Person} and \textit{Background}, two categories, and regress to a 4D vector of offsets to predict a more accurate bounding box. Finally, we use NMS to remove redundant boxes and set a threshold to filter out boxes with low confidence scores.  The final bounding boxes are regarded as actor spatial location, and they are also served for action recognition during inference.
\vspace{-3mm}

\paragraph{Temporal aggregation module.}
Above operations belong to a general image detection pipeline. However, our input is a video clip. Only using static images ignores temporal  context information and makes it hard to deal with challenging situations in videos, \emph{e.g.}, occlusion and motion blur. Therefore, we design a temporal aggregation module to enhance the  features of the keyframe proposals. Besides keyframe feature $X_k$, we select two frame features $X_{k-s}$ and $X_{k+s}$ with a distance of $s$ from the keyframe as reference frames. Both keyframe feature and reference frame features are fed into RPN and ROI pooling to generate keyframe proposal features and reference proposal features, notated as $F^k=\{f^k_1,f^k_2,...,f^k_N\}$ and $F^r=\{f^r_1,f^r_2,...,f^r_N\}$ respectively. Transformer is adopted here to aggregate features of reference proposals to generate more informative keyframe proposal. 
The transformer block is composed of self-attention layer and feed-forward network (FFN). The attention map produced by self-attention layer is computed by matching the transformed keyframe proposals $F^k$ (a.k.a. the queries) $Q=\phi(F^k)$ to 
another transformation of the reference proposals (a.k.a. the keys) $K=\theta(F^r)$, with $\phi$ and $\theta$ being learnable linear transformation. 
\begin{equation}
    A_d = \textrm{Softmax}(\frac{\phi(F^k)*\theta(F^r)^\top }{\sqrt{d}}),
\end{equation}
where $A_d\in \mathbb{R}^{N\times N}$ is the generated attention map, $d$ is the dimension for $F^k$ and $F^r$, and $*$ is the dot product. 
Considering that our goal is to enhance the features of the keyframe proposals with the reference proposals, we use the original reference proposal features as values instead of projecting them with a linear transformation. In the FFN layer, we apply a linear projection on the aggregated feature and add it to the keyframe proposals. Namely,
\begin{equation}
    {F^k}^{\prime} = F^k + \textrm{Linear}(A_d * F^r),
\end{equation}
where ${F^k}^{\prime}$ is the final aggregated proposal features, $\textrm{Linear}$ is the linear projection, and $*$ is the dot product.

\subsection{Action branch}
\label{3.3}

In the action branch, we emphasize on the variant interactions for action recognition. We propose a TransPC module to construct person-context interaction and it considers entity features, context and position information synchronously. Furthermore, we employ a memory feature bank to capture long-term temporal interaction.
\vspace{-2mm}

\paragraph{TransPC.}
In this module, we adopt transformer to integrate person feature and context feature to explore the interaction relationships. For a keyframe $\mathcal{F}_k$, its person bounding box set $P =\{p_1,p_2,...,p_n\}$ is obtained from our detection branch, where $n$ is the number of proposals. We aim to compute the correlation between each person  (target person $p^t_i\in P$) and other person (supporting person $p^s_j \in P$). We use align ROI pooling to crop 3D person feature from 3D feature map $X$. Then, the 3D person feature is converted to 1D person feature $f^t$ via temporal and spatial pooling. We obtain the sequence of person feature set $F^p=[f^p_1,f^p_2,...,f^p_n]$. Different from using $F^p$ as queries and keys in the conventional way, here we use new person-context features as keys to embed context features and position information. Specifically, for pairwise target person $p^t$ and supporting person $p^s$, we select a rectangle box which encloses two person bounding boxes with a minimum area as our interested region. The four coordinates of this new box, \iie, top-left and bottom-right coordinates $(x_1, y_1, x_2, y_2)$, are computed as:
\begin{eqnarray}
\begin{split}
    &x_1 = \textrm{min}(x^t_1, x^s_1),
    &y_1 = \textrm{min}(y^t_1, y^s_1),\\
    &x_2 = \textrm{max}(x^t_2, x^s_2),
    &y_2 = \textrm{max}(y^t_2, y^s_2),
\end{split}
\end{eqnarray}
where $(x^t_1, y^t_1, x^t_2, y^t_2)$ and $(x^s_1, y^s_1, x^s_2, y^s_2)$ are the coordinates of target person and supporting person, respectively.
This box keeps the most critical context and their relative position. We also use aligned ROI pooling following two convolution layers with zero padding to crop our person-context box from feature map $X$ and project it to generate the key and value. The convolution operation benefits retaining the spatial position and adding zero padding can further strengthen this effect. Finally, a max pooling is employed to transform it as a 1D feature $f^{pc}$. Because each target person $p^t\in P$ need to compute the $f^{pc}$ with each supporting person $p^s\in P$, our person-context feature set $F^{pc}$ is a $n \times n$ matrix but not a sequence. The $F^{pc}$ is defined as:
\begin{equation}
\begin{matrix}
F^{pc}=
\begin{pmatrix}
f_{11}^{pc} & f_{12}^{pc} & \cdots & f_{1n}^{pc} \\
f_{21}^{pc} & f_{22}^{pc} & \cdots & f_{2n}^{pc} \\
\vdots & \vdots & \ddots & \vdots \\
f_{n1}^{pc} & f_{n2}^{pc} & \cdots & f_{nn}^{pc} 
\end{pmatrix}.
\end{matrix}
\label{F_pc}
\end{equation}

The vanilla transformer deals with sequence key and value. We cannot directly adopt it to compute our attention map. To enable attention map calculation between our $F^p$ and $F^{pc}$, we first repeat the person feature sequence $F^p$ for $n$ times along row dimension to produce a $n\times n$ matrix ${F^{p}}^*$ which is represented as:
\begin{equation}
\begin{matrix}
{F^p}^* = 
\begin{pmatrix}
f_1^p & f_1^p & \cdots & f_1^p \\
f_2^p & f_2^p & \cdots & f_2^p \\
\vdots & \vdots & \ddots & \vdots \\
f_n^p & f_n^p & \cdots & f_n^p 
\end{pmatrix}.
\end{matrix}
\end{equation}
Then, We compute the Hadamard product of ${F^{p}}^*$ and $F^{pc}$ to obtain  attention map $A_a$:
\begin{equation}
\begin{matrix}
A_a = \sigma({F^p}^*\odot F^{pc}) \\
=\sigma(\begin{pmatrix}
f_1^p* f_{11}^{pc} & f_1^p* f_{12}^{pc} & \cdots & f_1^p* f_{1n}^{pc} \\
f_2^p* f_{21}^{pc} & f_2^p* f_{22}^{pc} & \cdots & f_2^p* f_{2n}^{pc} \\
\vdots & \vdots & \ddots & \vdots \\
f_n^p* f_{n1}^{pc} & f_n^p* f_{n2}^{pc} & \cdots & f_n^p* f_{nn}^{pc} 
\end{pmatrix}),
\end{matrix}
\label{A_a}
\end{equation}
where $\odot$ is Hadamard product, $\sigma$ is a softmax function, and $*$ is the dot product of two vectors.

The person feature aggregation is performed as a weighted summation of the person-context feature values with the attention map as  summation weights. Thus, we compute the Hadamard product of attention map $A_a$ and value matrix, and sum the output along the row dimension to generate the sequence of the aggregated features. Finally, we apply a residual connection to sum the person features and the aggregated features. Similar to \cite{DBLP:conf/eccv/TangXMPL20}, we also adopt a dense serial structure to integrate our TransPC blocks.
\vspace{-2mm}

\paragraph{Memory feature bank.}
Long-term memory features can provide effective temporal information to assist recognizing actions. Inspired by the Long-term Feature Bank (LFB) proposed in~\cite{Wu_2019_CVPR}, we build a memory feature bank to store both past and future person features for the long-term temporal interaction capture. During training, we store the person features according to the ground truth bounding boxes. During inference, we use the bounding boxes provided by our detection branch. We insert our memory feature bank after our TransPC module.

\section{Experiments on AVA}
The Atomic Visual Actions (AVA) \cite{DBLP:conf/cvpr/GuSRVPLVTRSSM18} dataset is collected for spatio-temporal action detection. In this dataset, each person on keyframes is annotated with a bounding box and corresponding action labels at 1 FPS. There are 80 atomic action categories including 14 pose categories and 66 interaction categries. This dataset contains 430 15-minute videos splitting into 235 training videos, 64 validation videos, and 131 test videos.

Since our method is designed for spatio-temporal action detection, we adopt AVA dataset as the main benchmark to conduct detailed ablation experiments. The results are evaluated with the official metric of frame level mean average precision (mAP) at spatial IoU $\ge 0.5$, and 60 categories with at least 25 instances in validation and test splits are used for evaluation following the conventional setup~\cite{DBLP:conf/cvpr/GuSRVPLVTRSSM18}.

\subsection{Implementation details}
\label{4.2}
\paragraph{Backbone.} 
With the modeling shift from CNNs to transformers in the vision community, pure transformer architectures have achieved top accuracy on the major video recognition benchmarks~\cite{DBLP:conf/icml/BertasiusWT21,DBLP:journals/corr/abs-2103-15691,DBLP:journals/corr/abs-2106-13230,DBLP:journals/corr/abs-2104-11227,DBLP:journals/corr/abs-2108-11575,DBLP:journals/corr/abs-2106-11297}. In this work, we adopt state-of-the-art Video Swin Transformer~\cite{DBLP:journals/corr/abs-2106-13230} as our backbone. Its base version (Swin-B) is selected, which consists of 4 stages, each containing 2, 2, 18, and 2 Swin Transformer Blocks. The original Swin-B performs $32\times$ spatial downsampling for the input videos. To maintain a larger spatial resolution of feature map $X$, we remove the last stage with the last patch merging layer to make the downsampling rate to be $16$. All other settings follow the recipe in~\cite{DBLP:journals/corr/abs-2106-13230}. Our backbone is pre-trained on Kinetics-600~\cite{carreira2017quo} dataset for action recognition task. To the best of our knowledge, this is the first work to explore the performance of transform-based backbone for this task. Besides the transformer-based backbone, we also use the 3D CNN backbone for fair comparison. We choose SlowFast~\cite{feichtenhofer2019slowfast} network with ResNet-101 structure which is pre-trained on Kinetics-700 dataset.
\vspace{-3mm}

\paragraph{Training and inference.}
The inputs of our network are 16 RGB frames, uniformly sampled from a 32-frame raw clip centered on a keyframe. All the video clips are scaled such that the shorter side becomes 256 and the longer side becomes 464, and then fed into backbone network initialized from Kinetics pre-trained weights. Random flipping is used during training. We train our network using the SGD optimizer with batch size 16. We train for 110k iterations with a base learning rate of 0.001, which is then decreased by a factor of 10 at 70k and 90k iteration. A linear warm-up scheduler~\cite{goyal2017accurate} is applied for the first 2k iterations. On AVA dataset, pose categories are mutually exclusive and interaction categories are not, so we use cross-entropy loss function for pose categories classification and binary cross-entropy loss function for interaction categories. To alleviate the deficiency of training data for person detector, we first use the data with ``person'' labels from MSCOCO~\cite{lin2014microsoft} to pre-train the detection branch. Since the data in MSCOCO are static images, the same images are stacked repeatedly to form video clips. During training, the person bounding boxes produced by RPN with IoU greater than 0.8 predicted will be fed into our action branch for action recognition. During inference, predicted person bounding boxes with a confidence score larger than 0.8 are used. In our memory feature bank, both 30 past and future clips are used. 

\subsection{Ablation study}
To verify the effectiveness of our method, we conduct ablation experiments on the validation set of AVA v2.2. The backbone we used is the modified Swin-B (more detail in Section \ref{4.2}).
\vspace{-3mm}

\paragraph{Decoupled \textit{vs}. coupled.}
In our method, we use a decoupled structure to decouple detection representation and action representation. Further, we experiment with decoupled structure and coupled structure. The results are shown in Table \ref{decouple}. The coupled structure refers to the structure of \cite{Girdhar_2019_CVPR}. After the RPN module generates person proposals, the detection branch integrates into the action branch. Both bounding box regression loss and action classification loss are upon the person features produced by the action branch. Our decoupled structure outperforms the coupled structure with a large margin, which demonstrates the effectiveness of our method.
\vspace{-3mm}

\paragraph{Detection branch.}
Our detection branch is responsible for generating person bounding boxes, which is the basis of our framework.
We evaluate our variants in our detection branch by ablating its temporal aggregation module (TAM) and MSCOCO data pre-training in Table \ref{detection branch}. \textit{Baseline} represents the model containing a basic faster-rcnn structure and a completed action branch. We can see that the temporal aggregation module can improve the results, which demonstrates that the temporal information is effective for our detection task but has been missed in previous methods. With more complicated video object detection methods~\cite{han2020exploiting,chen2020memory} having been developed, we believe that the performance of detection branch still has lots of  potential. One of the advantages of two-stage methods is that the off-the-shelf person detector can benefit from large-scale datasets in the person detection community. Similarly, in our one-stage method, extra pre-training data like MSCOCO dataset can also be employed, which brings in a significant gain. 
\vspace{-3mm}

\paragraph{The effectiveness of TransPC.}
Experiments in Table~\ref{TransPC} verify the effectiveness of our TransPC and compare TransPC with its counterpart method. In this table, \textit{Baseline} represents the model without TransPC module, and \textit{Person-person} represents the general person-person interaction module (similar to \cite{DBLP:conf/eccv/TangXMPL20}) without considering the context information. Our TransPC can 
largely enhance the mAP of \textit{Baseline} because the interaction relationships play a core role in this task. Our TransPC is also much better than the general person-person interaction method, demonstrating that the interaction relationship benefits from context information.

\begin{figure}[t]
\includegraphics[width=1.0\linewidth]{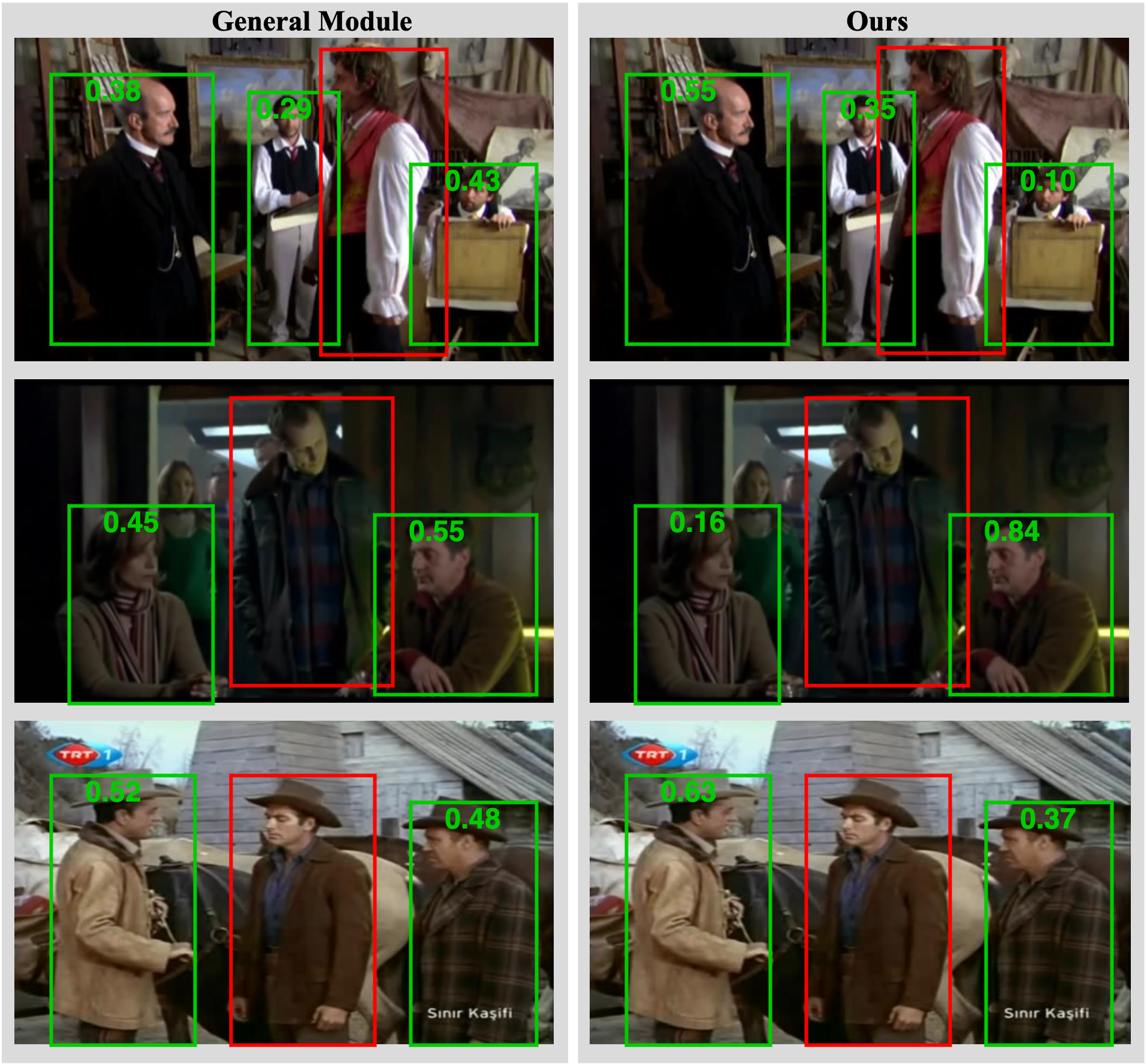}
\centering
\caption{We visualize attention weights from our TransPC and the general person-person interaction module. The actors of interest are in the red boxes and the other actors are in the green boxes. In each image, we remove the weight of actor of interest and re-normalize the rest attention weights to 1. }
\label{vis}
\vspace{-6mm}
\end{figure}

To further verify the effectiveness of our TransPC, we visualize attention weights from our TransPC and the general person-person interaction module in Figure \ref{vis}. The actors of interest are in the red boxes and the other actors are in the green boxes. In these multiple people cases, the general interaction module cannot distinguish the importance of other actors due to the lack of context and position information during building person-person relationship. In contrast, our TransPC can pay more attention to the actual supporting actors.
\vspace{-3mm}

\paragraph{Number of TransPC blocks.}
We adopt dense serial structure~\cite{DBLP:conf/eccv/TangXMPL20} to arrange our TransPC blocks. The effects of different numbers of TransPC blocks are shown in Table~\ref{number of blocks}. Considering the trade-off between accuracy and time consumption, we use three TransPC blocks in our method.
\vspace{-3mm}

\paragraph{Our TransPC \textit{vs.} other schemes.}
Furthermore, we explore different structures which could be an alternative to our TransPC, including: (i) directly use Eq.~\ref{F_pc} to generate attention map, (ii) use general person sequence feature as key to produce attention map and add the result of Eq.~\ref{F_pc}. Note that, only one block is used here for simplicity. Table~\ref{schemes} compares all these variants, with our choice outperforming other two variations.

\begin{table} [t]
\begin{center}
\small
\begin{tabular}{c|c}
\toprule
Structure & mAP  \\
\midrule
Coupled &  26.16\\
Decoupled (ours) & 28.82 \\
\bottomrule
\end{tabular}
\end{center}
\vspace{-5mm}
\caption{Performance evaluation on coupled structure and decoupled structure.}
\label{decouple}
\end{table}

\begin{table} [t]
\begin{center}
\small
\begin{tabular}{c|c}
\toprule
Variants of model & mAP  \\
\midrule
Baseline & 27.34 \\
Baseline + TAM & 27.75 \\
Baseline + TAM + MSCOCO & 28.82 \\
\bottomrule
\end{tabular}
\end{center}
\vspace{-5mm}
\caption{Performance evaluation on different components of detection branch.}
\label{detection branch}
\end{table}

\begin{table} [t]
\begin{center}
\small
\begin{tabular}{c|c}
\toprule
Variants of model & mAP  \\
\midrule
Baseline & 26.50 \\
Baseline + TransPC & 28.82 \\
Baseline + Person-person & 28.29 \\
\bottomrule
\end{tabular}
\end{center}
\vspace{-5mm}
\caption{Performance evaluation on the effectiveness of TransPC.}
\label{TransPC}
\end{table}

\begin{table} [t]
\begin{center}
\small
\begin{tabular}{c|c}
\toprule
Number of TransPC blocks & mAP  \\
\midrule
1 & 28.35 \\
2 & 28.68 \\
3 & 28.82 \\
4 & 28.84 \\
\bottomrule
\end{tabular}
\end{center}
\vspace{-5mm}
\caption{Performance evaluation on different number  of  TransPC  blocks.}
\vspace{-3mm}
\label{number of blocks}
\end{table}

\begin{table} [t]
\begin{center}
\small
\begin{tabular}{c|c}
\toprule
Scheme & mAP  \\
\midrule
Scheme i & 28.01 \\
Scheme ii & 28.10 \\
TransPC & 28.35 \\

\bottomrule
\end{tabular}
\end{center}
\vspace{-5mm}
\caption{Comparison of our TransPC with other schemes.}
\vspace{-1mm}
\label{schemes}
\end{table}


\subsection{Comparison with state of the art}
We compare our results with existing state-of-the-art one-stage and two-stage methods on the validation set of both AVA v2.1 (Table \ref{ava2.1}) and v2.2 (Table \ref{ava2.2}). For fair comparison, our experiments only use a single model and a single scale for testing. We provide results with SlowFast backbone and popular transformer-based Video Swin Transformer backbone. On both AVA v2.1 and v2.2, our results outperform all the results of one-stage methods by a large margin and are superior or comparable with two-stage methods, which indicates our method is effective. 
Comparing two backbones, Swin-B achieves a slightly better performance than SlowFast, which demonstrates that the transformer-based backbone is effective in this task. The per category results for our method are shown in our supplementary material.

As a one-stage method, another advantage of our method is the inference speed. We evaluate the average inference time of a single video clip on a typical two-stage method AIA~\cite{DBLP:conf/eccv/TangXMPL20} and ours in Table~\ref{speed}. Both use SlowFast backbone with ResNet-101 structure. Our inference time is only 64\% of that of AIA, showing its efficiency.

\begin{table} [t]
\begin{center}
\small
\begin{tabular}{c|c|c|c|c}
\toprule
Pipeline & Model & Modalities & Input & mAP  \\
\midrule
\multirow{6}{3em}{2-stage}
& ACRN~\cite{DBLP:conf/eccv/SunSVMSS18} & V+F & $32\times 2$ & 17.4 \\
& SlowFast~\cite{feichtenhofer2019slowfast} & V & $32\times 2$ & 27.3 \\
& LFB~\cite{Wu_2019_CVPR} & V & $32\times 2$ & 27.7 \\
& CA-RCNN~\cite{DBLP:conf/eccv/WuKWZW20} & V & $32\times 2$ & 28.8 \\
& AIA~\cite{DBLP:conf/eccv/TangXMPL20} & V & $32\times 2$ &31.2 \\
& ACAR-Net~\cite{DBLP:conf/cvpr/PanCSLS021} & V & $64\times 2$ & 30.0 \\
\midrule
\multirow{5}{3em}{1-stage}
& YOWO~\cite{kopuklu2019yowo}& V & $16\times 1$ & 19.2 \\
& Jiang et al.~\cite{jiang2018human}& V+F & $20\times$ -  & 21.7 \\
& VAT~\cite{Girdhar_2019_CVPR} & V & $64\times$ - & 25.0 \\
& Ours (SlowFast) & V & $16\times 2$ & 27.5 \\
& Ours (Swin-B) & V & $16\times 2$ & 27.7 \\
\bottomrule
\end{tabular}
\end{center}
\vspace{-5mm}
\caption{Comparison on AVA v2.1. V and F refer to visual frames and optical flow respectively. The input is shown as the frame number and corresponding sample rate.}
\vspace{-3mm}
\label{ava2.1}
\end{table}

\begin{table} [t]
\begin{center}
\small
\begin{tabular}{c|c|c|c|c}
\toprule
Pipeline & Model & Modalities & Input & mAP  \\
\midrule
\multirow{3}{3em}{2-stage}
& SlowFast~\cite{feichtenhofer2019slowfast} & V & $32\times 2$ & 29.0 \\
& AIA~\cite{DBLP:conf/eccv/TangXMPL20} & V & $32\times 2$ & 32.3 \\
& ACAR-Net~\cite{DBLP:conf/cvpr/PanCSLS021} & V & $64\times 2$ & 33.3 \\
\midrule
\multirow{3}{3em}{1-stage}
& YOWO~\cite{kopuklu2019yowo}& V & $16\times 1$ & 20.2 \\
& Ours (SlowFast) & V & $16\times 2$ & 28.5 \\
& Ours (Swin-B) & V & $16\times 2$ & 28.8 \\
\bottomrule
\end{tabular}
\end{center}
\vspace{-5mm}
\caption{Comparison on AVA v2.2. V and F refer to visual frames and optical flow respectively. The input is shown as the frame number and corresponding sample rate.}
\vspace{-1mm}
\label{ava2.2}
\end{table}

\begin{table} [t]
\begin{center}
\small
\begin{tabular}{c|c|c|c}
\toprule
Method & Detector & Action Recognition & Total time  \\
\midrule
AIA~\cite{DBLP:conf/eccv/TangXMPL20} & 0.106 & 0.156 & 0.262\\
\midrule
Ours & -& -& 0.168 \\
\bottomrule
\end{tabular}
\end{center}
\vspace{-5mm}
\caption{Comparison of the average inference time of each video clip (s/video clip) between two-stage method AIA and ours.}
\vspace{-2mm}
\label{speed}
\end{table}

\section{Experiments on UCF101-24}
\paragraph{Dataset.}
UCF101-24 is a subset of UCF101~\cite{soomro2012ucf101} consisting of 3207 videos with spatio-temporal annotations on 24 action categories. We conduct experiments on the first split of this dataset following previous methods. We use the corrected annotations provided in~\cite{singh2017online}.
\paragraph{Implementation Details.}
Following \cite{DBLP:conf/cvpr/PanCSLS021}, we use SlowFast with ResNet-50 structure as our backbone. The temporal sampling for the slow pathway is $8\times 4$, and the 32 frames as input are fed into the fast pathway. We pre-train it on the Kinetics-400 dataset. Other hyper-parameters are similar to the experiments on AVA.
\paragraph{Quantitative results.}
Table \ref{ucf} shows the result of UCF101-24 test set in frame-mAP with 0.5 IoU threshold. Our method surpasses another one-stage method with a considerable margin and is also competitive with two-stage methods. This outstanding performance illustrate the effectiveness and generality of our method again. We argue that UCF101-24 is not very suitable for most recent methods, including ours, in this task because the categories in this dataset are not interactive. Thus, many interaction relationships exploited by these methods are not very beneficial. Moreover, the quality of frames in this dataset is lower than AVA and MSCOCO, which adversely influences our detection results. 

\begin{table} [t]
\begin{center}
\small
\begin{tabular}{c|c|c|c}
\toprule
Pipeline & Model & Modalities & mAP  \\
\midrule
\multirow{7}{3em}{2-stage}
& T-CNN~\cite{hou2017tube} & V & 67.3 \\
& ACT~\cite{kalogeiton2017action} & V & 69.5 \\
& STEP~\cite{DBLP:conf/cvpr/YangY0XDK19} & V+F & 75.0 \\
& I3D~\cite{carreira2017quo} & V+F & 76.3 \\
& Zhang et al.~\cite{zhang2019structured} & V & 77.9 \\
& S3D-G~\cite{xie2018rethinking} & V+F & 78.8 \\
& AIA~\cite{DBLP:conf/eccv/TangXMPL20} & V & 76.8 \\
\midrule
\multirow{2}{3em}{1-stage}
& YOWO~\cite{kopuklu2019yowo}& V & 70.5 \\
& Ours & V & 74.8 \\
\bottomrule
\end{tabular}
\end{center}
\vspace{-5mm}
\caption{Comparison on UCF101-24 split 1. V and F refer to visual frames and optical flow respectively. The metric we used is frame-mAP.}
\vspace{-5mm}
\label{ucf}
\end{table}

\section{Limitations}
There are several limitations of this work. First, the proposed DOAD method is only evaluated on the commonly used AVA and UCF101-24 datasets. More evaluations on other datasets will be better. Second, limited by the more GPU memory consumption of our key matrix, our method is difficult to be applied in crowded scenes.

\section{Conclusion}
In this paper, we propose a new effective and efficient one-stage sptio-temporal action detection network, DOAD. We decouple the person detection and action recognition into two branches to alleviate the issue of different optimal context supports. Moreover, different from independently utilizing kinds of context, we present a novel TransPC module to integrate the person and context features to capture the interaction relationships. Our method significantly outperforms all the existing one-stage methods and is superior or comparable with two-stage methods on challenging benchmarks. Our method provides a new and strong one-stage framework which still has tremendous potential. 
In the future, we plan to further study how to improve the performance of person detection and capture more fine-grained detail features for action recognition.
{\small

\bibliographystyle{IEEEtran}
}

\end{document}


\title{Supplementary for Decoupled One Stage Action Detection Network}

\author{First Author\\
Institution1\\
Institution1 address\\
{\tt\small firstauthor@i1.org}
\and
Second Author\\
Institution2\\
First line of institution2 address\\
{\tt\small secondauthor@i2.org}
}

\maketitle


\section{Additional experiments}

\subsection{Detection performance (action agnostic)}
As a one-stage method, we use the person bounding boxes generated by our network instead of an off-the-shelf detector. We now investigate in detail at the detection performance. We report the performance in Table~\ref{detection} with a standard 0.5 IOU threshold. The off-the-shelf detector is Faster R-CNN framework with ResNeXt-101-FPN backbone from maskrcnn-benchmark, which is widely applied in two-stage methods [\textcolor{green}{48}, \textcolor{green}{38}, \textcolor{green}{26}]. The model is first pre-trained on ImageNet, then fine-tuned on MSCOCO dataset, and finally fine-tuned on AVA dataset for higher person detection precision. We can see that our person detection result is still lower than the specialized detector, which is the crucial reason that the performance of one-stage methods cannot surpass the state-of-the-art two-stage methods.

\subsection{Backbone modification}
We are the first to use transformer-based backbone, Swin-B, in this task. A tough nut is how to maintain a larger spatial resolution of the feature map due to the overlarge spatial downsampling rate in the original version. The downsampling of Swin-B is mainly from the patch merging layer followed by a swin transformer block. We propose two schemes to modify the Swin-B: (i) removing the last patch merging layer and its following swin transformer block; (ii) just removing the last patch merging layer and modifying the dimension of the weights of the last swin transformer block. Note that the last swin transformer block in scheme (ii) cannot be initialized from a pre-trained model and can only be trained from scratch. Their results are presented in Table \ref{backbone}. Scheme (i) is slightly higher than scheme (ii) and contains fewer parameters. Thus, we adopt scheme (i) in our method.

\section{Qualitative results}
\begin{figure}[t]
\includegraphics[width=1.0\linewidth]{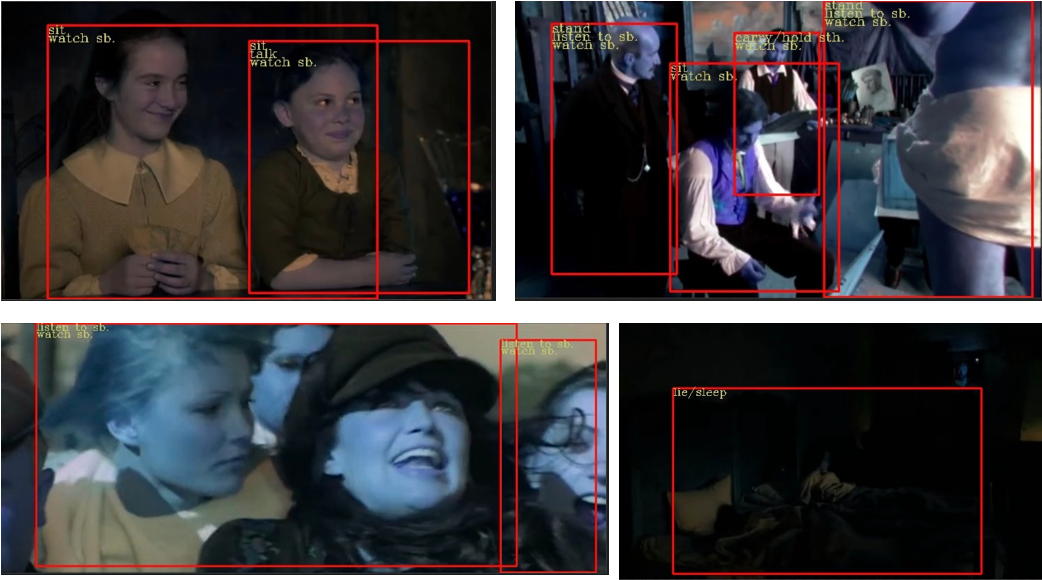}
\centering
\caption{Per category results for the proposed network and the baseline model on the validation set of AVA dataset.}
\label{qualitative}
\end{figure}

We randomly visualize some cases of our model in Figure \ref{qualitative}. Our model is able to exploit the person-context relationships to recognize interaction categories such as ``watch sb" and ``listen to sb", which are inherently hard if only focus on the actor, as shown in the first row of Figure \ref{qualitative}. There are two failure detection cases in the second row, which shows that our detection is hard to handle crowded and dark scenes.

\begin{table} [t]
\begin{center}
\small
\begin{tabular}{c|c}
\toprule
Method & mAP (IOU@0.5)  \\
\midrule
Off-the-shelf detector & 93.5 \\
Ours & 89.9 \\
\bottomrule
\end{tabular}
\end{center}
\caption{We perform classification-agnostic evaluation to evaluate the performance of our person detection and compare it with the off-the-shelf detector.}
\label{detection}
\end{table}

\begin{table} [t]
\begin{center}
\small
\begin{tabular}{c|c}
\toprule
Scheme & mAP (IOU@0.5)  \\
\midrule
 i & 28.8 \\
 ii & 28.5 \\
\bottomrule
\end{tabular}
\end{center}
\vspace{-5mm}
\caption{Comparison of backbone modification schemes.}
\vspace{-3mm}
\label{backbone}
\end{table}

\begin{figure*}[!t]
\includegraphics[width=0.95\linewidth]{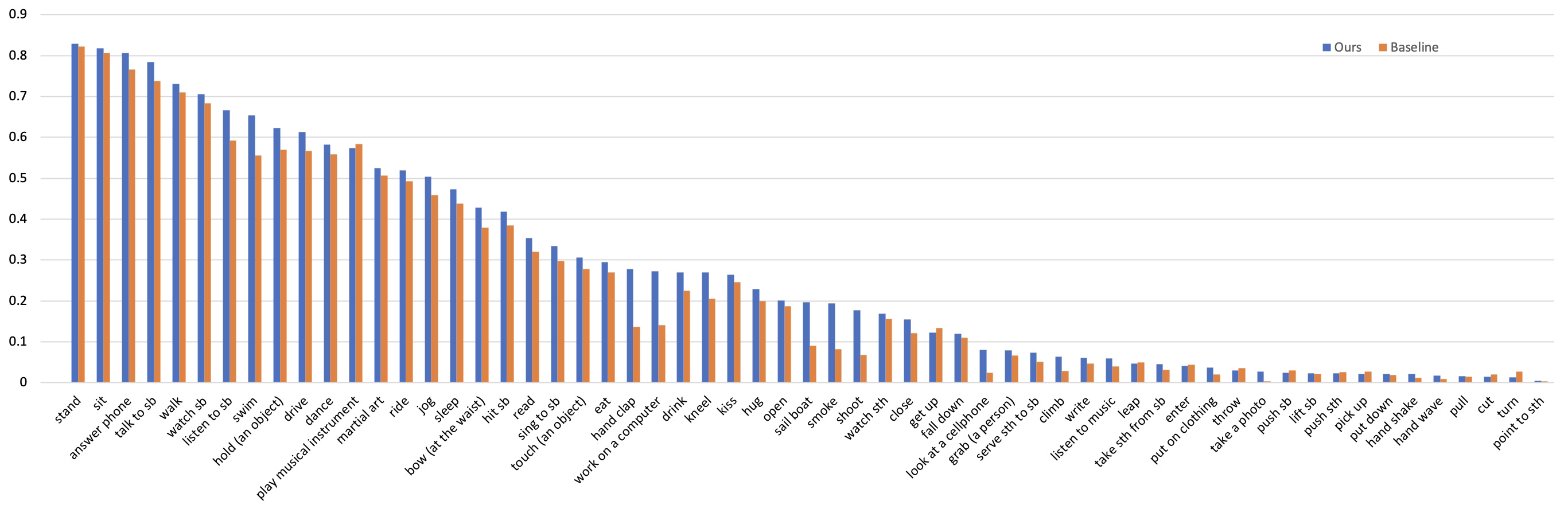}
\centering
\caption{Per category results for the proposed network and the baseline model on the validation set of AVA dataset.}
\label{visualization}
\end{figure*}

\section{Per category analysis}
The per category results for our method and the baseline (the full model without TransPC) are shown in Figure \ref{visualization}. Our method improves the baseline performance in about 50 out of 60 classes. We can see that the categories getting the largest performance boost are from interaction categories, e.g., ``hand clap", ``work on a computer", ``smoke", and ``listen to sb", which require more attention on the supporting actors and context.
